\documentclass{article}
\usepackage{ja_eng}
\usepackage[dvips]{graphicx}
\usepackage[ansinew]{inputenc}

\title{ViT Cane: Visual Assistant for the Visually Impaired }

\author{Bhavesh Kumar\\ Bhavkumar21@gmail.com \\}

\begin{document}

\maketitle

\begin{abstract}
{\em Blind and visually challenged face multiple issues with navigating the world independently. Some of these challenges include finding the shortest path to a destination and detecting obstacles from a distance. To tackle this issue, this paper proposes ViT Cane, which leverages a vision transformer model in order to detect obstacles in real-time. Our entire system consists of a Pi Camera Module v2, Raspberry Pi 4B with 8GB Ram and 4 motors. Based on tactile input using the 4 motors, the obstacle detection model is highly efficient in helping visually impaired navigate unknown terrain and is designed to be easily reproduced. The paper discusses the utility of a Visual Transformer model in comparison to other CNN based models for this specific application. Through rigorous testing, the proposed obstacle detection model has achieved higher performance on the Common Object in Context (COCO) data set than its CNN counterpart. Comprehensive field tests were conducted to verify the effectiveness of our system for holistic indoor understanding and obstacle avoidance.
}

\end{abstract}

\section{INTRODUCTION}
For sighted people, when they enter an unfamiliar indoor environment, they can observe and perceive the surrounding environments through their vision. However, such a global scene understanding is challenging for people with visual
impairments. They often need to approach and touch the objects in the room one by one to distinguish their categories and get familiar with their locations. This is not only inconvenient but also creates some risks for visually impaired
people in their everyday living and travelling tasks. In this
work, we develop a system to help vision-impaired people
understand unfamiliar indoor scenes. \\
\subsection{Related Work}
Several products have tried solving the problem of obstacle detection and world navigation. For example, wearable devices that contain sensors, like an RGB camera, Lidar or sonar sensors, to understand the surrounding scene. In most cases the visual assistant relays data using audio feed backs or tactile sensors. These, vision-based assistive systems employ deep learning architecture to improve perception tasks. The assistive system proposed by [3] learns from RGBD data and predicts semantic maps to support the obstacle avoidance task.[4] integrated sensor based, computer vision-based, and fuzzy logic techniques to detect objects for collision avoidance. \\

A CNN-based framework DEEP-SEE[5], integrated into a novel assistive device, was designed to recognize objects in the outdoor environment. In [6], a kinetic real-time CNN was customized for the recognition of road barriers to support navigation assistance, which are usually set at the gate of a residential area or working area. The wearable system [7] with a pair of smart glasses informs the visually impaired people based on semantic segmentation and 3D reconstruction. In [8], a wearable belt is developed with deep learning system to pinpoint the exact locations of surrounding objects and the scene type in real-time. \\

Differently, our work focuses on applying a vision transformer to apply the same techniques in an efficient manner requiring much less computational resources. The ViT Cane is in the style of an actual cane and is uses a simple RGB camera and 3 motors to relay tactile information to the user.
\section{Methods}
We considered two approaches: one using YOLOv3 and the second using Vision Transformers. These are described below.
\subsection{Method 1: YOLO Architecture}
\begin{figure}[htbp]
\centerline{\includegraphics[width=8cm,angle=0]{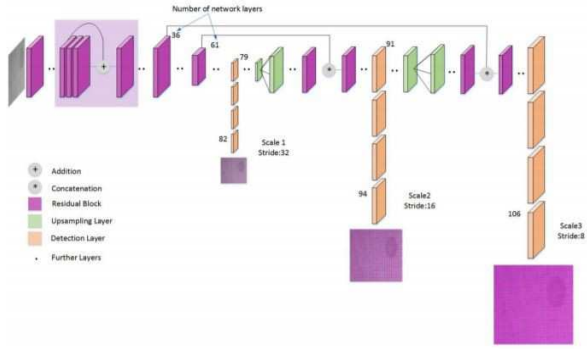}}
\caption{YOLO Architecture (Citation [2])} \label{fig:2}
\end{figure}
The first prototype of this solution is designed using the simple YOLOv3 model (refer to figure 1) to identify obstacles in the users path. We trained this model using the COCO data set to identify objects in the path of the user and relay that information using audio output. The novelty of the system relies in YOLO's speed and it's increased accuracy with the large COCO data set. This implementation lacks space efficiency which proves to be a problem when transferring the model into a Raspberry Pi to assemble the cane thus after research we decided to use the Vision Transformer model instead. \\

\begin{figure}[htbp]
\centerline{\includegraphics[width=8cm,angle=0]{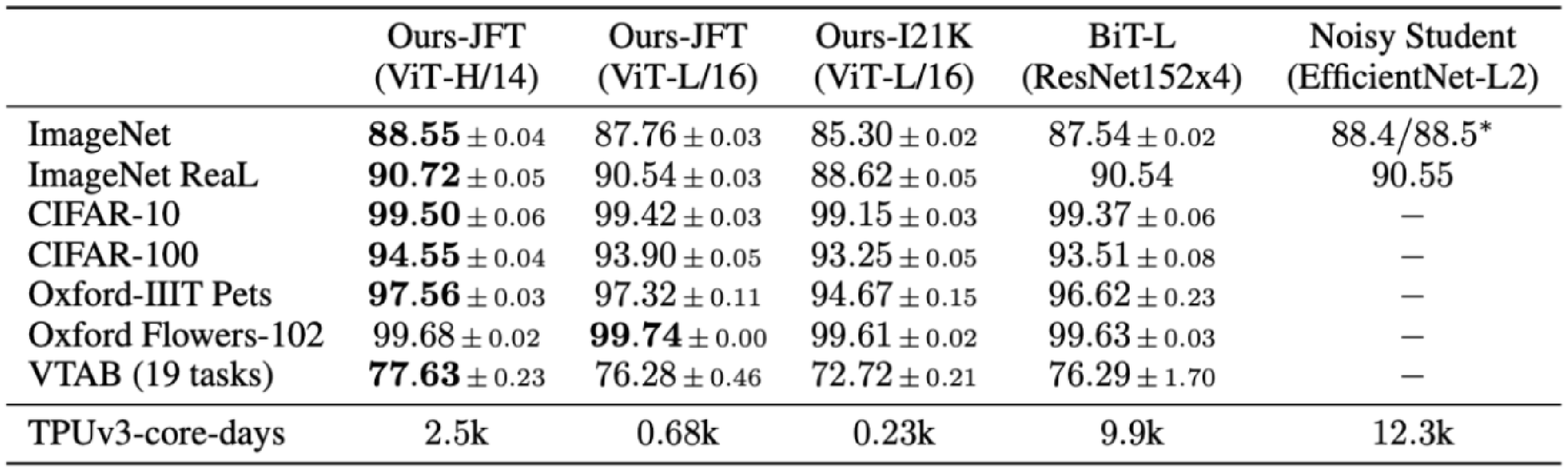}}
\caption{ViT compared to CNN (Citation [1])} \label{fig:3}
\end{figure}
In a prior work, researchers have demonstrated that Vision Transformers tend to outperform CNN models in object detection [1]. ViT requires much less computational resources than the other 2 CNN models. Even in our COCO dataset Vision transformers are more efficient compared to the old CNN obstacle detection model. Thus, in our cane, we will be utilizing the Vision transformer model to detect obstacles.\\

\subsection{Method 2: Vision Transformers}
The ingenuity of this paper relies on the utilization of the recent vision transformers model to detect object. In this section we'll go over the architecture of the Vision Transformer proposed by [1]  The models starts by splitting an image into 9 patches without overlap. Each patch is vectorized using its RGB channels. These starting vectors (x vectors) are transformed through a dense layer which is essentially a linear transformation layer with trained weights and biases. Through the dense layer, we get transformed vectors we will call z vector which is assigned a positional encoding based on where each vector was positioned in the image. This positional encoding is an important part of our transformed vector, as the positional encoding allows the transformer to analyze the image with all the patches in the right order. At this point, a classification token or CLS token (x vector) is established in our transformer. The CLS token is put into an embedded layer to give us another transformed vector with the same shape as the z vectors. Now all of the z vectors are put into a series multi-Head Self-Attention and Dense layer which is commonly known as Transformer Encoded Network. Through this, we get classification vectors. Of all the classification vectors, the one that originated from the CLS token is the most important as it is aware of the image as a whole. We feed this vector into a softmax classifier to get the classification of an object. \\

\begin{figure}[htbp]
\centerline{\includegraphics[width=8cm,angle=0]{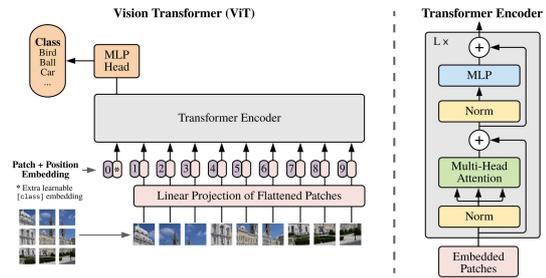}}
\caption{Vision Transformer Architecture (Citation [1])} \label{fig:2}
\end{figure}

\section{IMPLEMENTATION OF ViT CANE}
\begin{figure}[htbp]
\centerline{\includegraphics[width=7cm,angle=0]{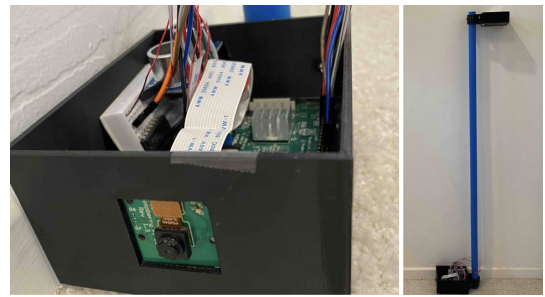}}
\caption{ViT Cane} \label{fig:4}
\end{figure}
Using the Vision Transformer model, we are able to identify objects in the surroundings but a major inefficiency arises that not all objects are obstacle. In fact, most of them are unnecessary until unless the user is standing close to them. Thus, ViT Cane runs an obstacle detection model to categorize objects as path obstacles. In real time, we separate the live feed frame by frame which is sent to the vision transformer model to get object detection. For each frame, the positions of the bounding boxes are saved which are later compared with to the following frame. Through this comparison, we are to determine obstacles in the path of the user. We match the coordinates of the every object in one frame to another to get a rough estimate of the continuously moving object. We then compare the areas of the bounding box of the object over time, concluding a generally increasing bounding box as an obstacle moving towards the cane as shown in figure. These obstacles are then separated based on their placement in each frame and then transcribed to the motors to represent obstacles. \\

The ViT Cane is a minimally designed smart cane created to be replicated at the cheapest cost possible. The cane has two 3D printed parts one for housing the major electronic components like the Raspberry Pi, Camera and the micro controller, the other is used to store the motors. There are a total of 3 motors representing right, front, left respectively for the user to understand which direction the obstacle lies. These parts were custom designed but could be replicated using any two box with enough space to fit the electronics and the motors. These two parts are connected through a 4ft PVC which with holes on each end to run the wire for the motors. The entirety of the cane costs about 61 USD and can be assembled at home.
\\

\section{RESULTS}
\begin{tabular}{ |p{1.5cm}|p{1.5cm}|p{1.5cm}|p{1.5cm}|  }
 \hline
 \multicolumn{4}{|c|}{Average Time Taken Per Course} \\
 \hline
& White Cane & YOLO Cane & ViT Cane\\
 \hline
 Course 1 & 42 sec & 63 sec & 57 sec\\
 Course 2 & 88 sec & 92 sec & 85 sec \\
 Course 3 & 119 sec & 124 sec & 110 sec \\
 Course 4  & 158 sec & 160 sec & 156 sec\\
 Course 5 & 193 sec & 201 sec & 187 sec\\
 \hline
\end{tabular} \\

Table 1 displays data of the user study that we performed. I compared ViT cane against the YOLO Cane and a White Cane to compare the speed and accuracy of the three different assistive methods. I designed 5 unique obstacle courses meant to mirror various day to day situations for a visually impaired person. Each of the 5 obstacles courses are detailed below.
\begin{itemize}
    \item Course 1: 50 meter long with 6 obstacles
    \item Course 2: 100 meter long with 8 obstacles
    \item Course 3: 150 meter long with 10 obstacles
    \item Course 4: 200 meter long with 12 obstacles
    \item Course 5: 250 meter long with 14 obstacles
\end{itemize}
These 5 courses were tested by 5 volunteers who were blindfolded and asked to navigate the courses with their respective assistive devices. The obstacles on the course are common item on the road. Some of the items used as obstacles include: trashcans, street signs, bicycles and traffic cones. \\ 

As the table suggests, the ViT is time efficient giving the user an early enough heads up for them to avoid stationary obstacles. The cane performs significantly better than the YOLO Cane as described in method 1 and is comparable to the White Cane. In addition to the ViT Cane being time efficient, it also serves to be more accurate in obstacle detection compared to the YOLO cane. On all the courses, ViT Cane users had only 1 obstacle collision compared to YOLO Cane users who had on average 1 obstacle collision per course. According to the user, the YOLO Cane was slow in warning users of the obstacles which caused delay in course and increased collision risk. The users of the study also reported the White Cane to be more intuitive but explained that the ViT Cane and the YOLO Cane were more informational about the surrounding obstacles which makes them a better use once you get thoroughly accustomed to it.  

\section{FUTURE}
Overall, the ViT Cane serves to be an efficient alternative for the visually impaired in some cases but it does have it's shortcomings. While creating the obstacle course, I realized that there is a possibilities for obstacles to be above the user. In this case, you can argue that even the established White Cane cannot detect the obstacle but since this is an advanced cane I would like to create an overhead feature for the ViT Cane. \\
The Cane also does not have the capability to be used on real roads due to the fact that cars and other automobiles are way too fast for the ViT model to detect and relay to the user. The lack of roadside assistance by the ViT Cane is a major flaw. In the future, I'd like to devise a more time efficient method for the ViT Cane to process obstacles at a faster rate so that it could be used everywhere. This work can be further researched through the implementation of better hardware and it's impact on the efficiency of the cane. 

\section{CONCLUSION}
In this paper I have designed a competitively effective solution for obstacle detection. The proposed solution: ViT Cane utilizes the recent Vision Transformer architecture to detect obstacles in a fast and accurate manner. The objects detected are then classified as obstacles and eventually relayed to the user through motors integrated at the handle of the cane. The experiment demonstrates the efficiency of the ViT Cane in real-world scenarios captured by the five courses.

\section*{ACKNOWLEDGEMENT}

This work is supported by SupportVectors, CA. In particular, I appreciate the valuable mentorship provided to me by Dr. Asif Qamar. \\
Thanks Mrs. Kristin Berbawy and Berbawy Makers for helping me print the housing units for this project.

\end{document}